# Human Abnormality Detection Based on Bengali Text

M. F. Mridha
Department of Computer Science & Engineering
Bangladesh University of Business & Technology
Dhaka, Bangladesh
firoz@bubt.edu.bd

Md. Saifur Rahman
Department of Computer Science & Engineering
Bangladesh University of Business & Technology
Dhaka, Bangladesh
saifurs@gmail.com

Abu Quwsar Ohi
Department of Computer Science & Engineering
Bangladesh University of Business & Technology
Dhaka, Bangladesh
quwsarohi@gmail.com

*Abstract*— In the field of natural language processing and human-computer interaction, human attitudes and sentiments have attracted the researchers. However, in the field of human-computer interaction, human abnormality detection has not been investigated extensively and most works depend on image-based information. In natural language processing, effective meaning can potentially convey by all words. Each word may bring out difficult encounters because of their semantic connection with ideas or categories. In this paper, an efficient and effective human abnormality detection model is introduced, that only uses Bengali text. This proposed model can recognize whether the person is in a normal or abnormal state by analyzing their typed Bengali text. To the best of our knowledge, this is the first attempt in developing a text based human abnormality detection system. We have created our Bengali dataset (contains 2000 sentences) that is generated by voluntary conversations. We have performed the comparative analysis by using Naïve Bayes and Support Vector Machine as classifiers. Two different feature extraction techniques count vector, and TF-IDF is used to experiment on our constructed dataset. We have achieved a maximum 89% accuracy and 92% F1-score with our constructed dataset in our experiment.

## I. Introduction

At present, the internet is the most important part of our daily life, where we express our emotions, judgments, appreciations, etc. We text each other, comment on different online products, comment on another person's comment, provide opinions and most of them are happening in text form. So text is a great source for research that can be used to identify human characteristics.

There exists a strong relationship between human behavior, sentiment, and abnormality. Although, the terms human behavior, and sentiment relate to each other. Behavior is a human attitude that is the physiological activity or feeling performed towards some evaluation [1]. Repeatedly, the human sentiment defines the same characteristics, and in the largest sense, it is expressed in written words or speech. On the contrary, human abnormality explains the variation from absolute mental health. Human sentiment is the closest expression to human abnormality. Human abnormality mostly defines negative-sentiments, although a negative sentiment may or may not be defined as a human abnormality. Table I contains explanations that target to the relation between human abnormality and sentiment. By mental condition, two states, normal and abnormal is represented which is our fundamental finding.

The human abnormality detection problem is closely related to human sentiment analysis in the text. Sentiment analysis is the process of extracting emotions, opinions or a process of examining the attitude or behavior of humans. After examination analysis provides review based that is based on normal or abnormal behavior. As a result, an abnormality detection problem can be verified as a binary classification problem.

TABLE I.  Difference Between Sentiment and Mental Condition

| Expression | Sentiment | Mental Condition |
|---|---|---|
| সে আমায় ভালোবাসেনি | Negative | Normal |
| আজ কে যদি আমার মৃত্যু হইতো | Negative | Abnormal |
| আমি এটা পারবো | Positive | Normal |

The Bengali language is one of the most spoken languages in the world. People are using the Bengali language to express their opinions, emotions, etc. on online blogs, social media sites. Caused by this, the Bengali language is a great field for research.

A human sentiment can express love, sorrow, anger, depression, happiness, judgments, decision, suicidal, etc. Among all these types of sentiments, some special type of sentiment can be defined as an abnormal sentiment. Extracting or targeting abnormality from textual data is challenging yet not impossible.

Human abnormality can become a broad field of study, as it targets the mental balance of human characteristics. Abnormality detection based on the text has much potential, which can be utilized to detect abnormal state detection based on online messaging systems. An abnormality detection system can easily filter abnormal messages from online messaging systems, which may become invaluable in the reduction of online criminal/abnormal activities.

In this paper, we extract information from our text by which we will classify where a sentence is containing abnormal or normal attitudes through Machine Learning (ML). Along with the implementation of text-based abnormality detection, the contribution of this paper also includes the distinction between abnormality and negative sentiment states.

---

The rest of the paper is organized as follows. Section II outlines the related works in different languages. Section III presents the methodology of the proposed architecture. We present the result analysis in Section IV. Lastly, future work and conclusion are presented in Section V.

## II. RELATED WORK

In recent times, the classification of emotions into text is popular with the Natural Language Processing (NLP) researchers. Though it is harder to find exact research on human abnormality detection from the text, few works are done related to this topic, which is done for the English language and mostly links to attitude analysis as well.

To analyze the sentiment polarity from the text, rule-based linguistic models are created [2], [3], [4]. ML approach is popular to analyze sentiment from text, images, etc. Go, Bhayani and Huang used emoticons in the training corpus [5]. SVM, Naïve Bayes, and MaxEnt are used as a classifier. Davidiv, Tsur, and Rappoport used emoticons and hashtags to recognize the sentiment label and using the KNN algorithm to train a supervised sentiment classifier [6]. Pak proposed a method which consists of Naïve Bayes classifiers with POS-tag and n-gram features [7]. Alena, Helmet and Mistru design a system to analysis textual attitude [15]. It is based on compositionality principles and rules for semantically distinct verb classes. Balahur et al. do his research on different languages, Spanish, Germany, French with a Machine translation system [8]. SVM is used in the training phase. Kaur et al. proposed an algorithm with a unigram and simple scoring method for sentiment analysis from Punjabi text [9].

Attitude analysis with Bengali text is first work though sentiment analysis is studied in various topics. Various methods are explored to achieve the goal. Hasan et al. [10] gathered and combine the sentiment orientation for each sentence to recognize sentiment on Bengali text. They put the phrase patterns to match with predefined phrase patterns. The semi-supervised bootstrapping method is used for sentiment extraction from text with two labels of polarity positive and negative by Chowdhury et al. [11]. Nabi et al. represented a method of TF-IDF to extract sentiment [12]. They ignored mixed sentences and a few noisy data in their system. Shaika et al. identify sentiment with negative and positive polarity [13]. They used a semi-supervised bootstrapping approach to develop the training corpus and SVM and Maximum Entropy for classification. Tabassum and Khan used the random forest to classify sentence sentiment into positive and negative [14].

All the research works that are conducted in extracting sentiment rather than detecting abnormality of texts. In our work, we have classified human behavior as normal or abnormal using our own built corpus. We carried out different feature extraction and classification methods to implement the best possible architecture as well.

## III. METHODOLOGY

This method identifies human abnormality detection based on Bengali Text. This method identifies two states using text analysis, whether a person is normal or abnormal. To do this, first, we extract features from the text, then we use a classifier to classify sentences into normal and abnormal. The complete methodology is divided into the following steps: (a) Data collections and Preprocessing, (b) Features Extraction, and (c) Classification. Fig. 1. represents a visual of the mentioned work flow.

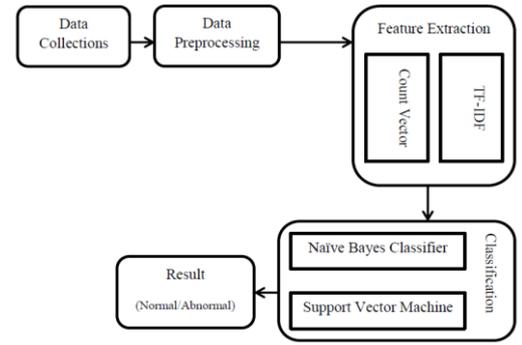

Fig.1. Work Flow of Human Attitude Analysis

### A. Data Collections and Preprocessing

In machine learning, data is the power of any model. Since there was no work on this very topic in the Bengali language, we had to create our very first human abnormality dataset based on Bengali Text. The data was gathered from volunteers who made conversations on social media sites like Twitter, Facebook, WhatsApp, Messenger, etc. The dataset was collected in the Bengali Text form.

We need to preprocess the data because we collected data from social media. The sentences of the dataset had many irrelevant tags, spaces, emojis, etc. which was removed by an automated script. The dataset contained sentences with appropriate labels (normal/abnormal). The sentences of the dataset were appropriately classified by multiple specialists. Each of the data is a Bengali sentence that contains two target values, 1 (abnormal) or 0 (normal). Table II and II contain an illustration of some normal and abnormal sentences of the dataset. The dataset contains 2000 sentences, where 814 are the abnormal sentence, and the rest of the sentence is the normal sentence.

TABLE II. NORMAL ATTITUDE

| Normal Attitude |
|---|
| মা বাবা বেঁচে থাকা অবস্থায় তাদেরকে কখনও স্বপ্নে দেখিনি |
| আমি যদি আমার অফিস টাইমের ফাকে নিয়মিত ওর খোঁজ নিতাম |
| সে একটা সুযোগ চায় |
| আমি পারবো কিভাবে |

TABLE III. ABNORMAL ATTITUDE

| Abnormal Attitude |
|---|
| আজ কে যদি আমার মৃত্যু হইতো |
| আমি ওকে মেরে ফেলব |
| মানুষ হতে চাও মদ খাও |

### B. Features Extraction

Features extraction indicates information gathering from the text in NLP. A feature uniquely describes the data by its properties. In ML, feature extraction also recognized as a dimensionality reduction technique. There exist many methods to extract features from the text. We will use two different techniques, Count Vector and TF-IDF Vector. We experiment with these two different feature extraction techniques and evaluate their accuracy in Section IV.

Word embedding is the numerical representation of text. Frequency-based and prediction based word embedding are the popular categories of word embedding. Count Vector and TF-IDF Vector are the subset of frequency-based word embedding.

*1) Count Vector*

Count Vector preserves the number of occurrences of the words of a text document. In Count Vector, a dictionary is constructed that comprises all unique data from all documents. Then it encodes a sentence based on the dictionary, where the encoded data holds the number of appearances of a word in that sentence.

*2) TF-IDF*

TF-IDF consists of two terms, TF which stands for Term Frequency and IDF stands for Inverse Document Frequency. TF-IDF presents the correlated importance of terms into a document and the whole corpus. The following formula give us TF-IDF score for a term T,

$$TF(T) = \frac{\text{number of occurrences of T in a document}}{\text{total number of T from that document}} \quad (1)$$

$$IDF(T) = \log_e \frac{\text{total number of documents}}{\text{number of documents which contains T}} \quad (2)$$

$$TF\text{-}IDF = TF \times IDF \quad (3)$$

Here, term T could be character, word, or n-gram. We use a word as a token or term to generate TF-IDF into our experiment.

*C. Classification*

The scope of the paper relates to supervised classification methods. A supervised classifier is a kind of model which is trained with training dataset with features and correct target classes. The features of the dataset are extracted through the feature extraction techniques, Count Vector and TF-IDF, which are elaborated in Section B. This section refines on the methods that were performed to build classifiers that perform abnormality detection.

*1) Naïve Bayes*

Naïve Bayes classifier is based on Bayes's theorem with the naïve assumption that the presence of a feature in a class is independent of the presence of any other features.

Assume $T = \{t_1, t_2, ..., t_n\}$ is the dependent feature vector with given a class $c$,

$$P(c|t_1,t_2,...,t_n) = \frac{P(c)P(t_1,t_2,...,t_n|c)}{P(t_1,t_2,...,t_n)} \quad (4)$$

Using naïve assumption, we can write (4) as

$$P(c|t_1,t_2,...,t_n) = \frac{P(c)\prod_{i=1}^{n} P(t_i|c)}{P(t_1,t_2,...,t_n)} \quad (5)$$

$$P(c|t_1,t_2,...,t_n) = P(c)\prod_{i=1}^{n} P(t_i|c) \quad (6)$$
[Since, $P(t_1,t_2,...,t_n)$ is constant]

$$\hat{c} = \text{argmax}\left(P(c)\prod_{i=1}^{n} P(t_i|c)\right) \quad (7)$$

Equation (7) is used to find expected class with maximum probability.

There exist several varieties of Naïve Bayes classifiers among which the most used classifiers are, Gaussian Naïve, Multinomial Naïve, and Bernoulli Naïve. We select Multinomial Naïve for our classification because it is most suitable to process text-based data. We used Count Vectors and TF-IDF as features that are discrete values and, Multinomial Naïve works on multinomial distributed data. Smoothing is used in Multinomial Naïve using the following equation,

$$\hat{\theta}_{ci} = \frac{N_{ci} + \alpha}{N_c + \alpha n} \quad (8)$$

In the (8) $N_{ci} = \sum_{t \in TS} t_i$ is the number of times feature $i$ appears in a sample of class $C$ in the training set $TS$, and $N_y = \sum_{i}^{n} N_{ci}$ is the total count of all features for class $c$. $\alpha$ is smoothing prior, for Laplace smoothing $\alpha=1$ and $\alpha<1$ called Lidstone smoothing.

*2) Support Vector Machine:*

SVM is a well-known machine learning algorithm to classify data that is more comfortable to use than the neural network and receives good accuracy as well. The principle of SVM is separating data into different classes by finding the optimal hyperplane, which has a maximum margin with minimum error. Following steps will describe the mathematics behind SVM,

Let's assume a training set TS,

$$TS = \{(x_1,y_1), (x_2,y_2), ..., (x_n,y_n)\} \quad (9)$$

$$TS = \{x_i, y_i\}_i^n; 1 \leq i \leq n \quad (10)$$

Where in (10) $x_i \in R^d$ is the input vector for the i[th] training data and $y_i \in (1,-1)$ is class label for i[th] training data and n is the number of training data set. The SVM follow the following decision function

$$y = \text{sign}\left(\sum_{i=1}^{n} y_i \alpha_i K(x,x_i) + \beta\right) \quad (11)$$

Here in (11) K is the kernel function which explained later, $\alpha$ and $\beta$ are the parameters where $\alpha = \{\alpha_1, \alpha_2, ..., \alpha_n\}$. To train the SVM, we should find $\alpha$ that minimize the objective function:

$$\min \frac{1}{2} \sum_{i=1}^{n} \sum_{j=1}^{n} \alpha_i \alpha_j y_i y_j K(x_i,x_j) - \sum_{j=1}^{n} \alpha_j \quad (12)$$

Equation (12) Subject to the constrained $\sum_{i=1}^{n} \alpha_i y_i = 0$, $0 \leq \alpha_i \leq C$.

So for training an SVM, we need to solve the quadric programming optimization problem with n number of parameters.

A kernel function is a process to represents relations between two data points in the feature space. It helps to minimize the complexity of finding the mapping function. Kernel function has the following categories,

- The linear $K(x,x_i) = x^{TS} x_i$
- Polynomial $K(x,x_i) = (\gamma x^{TS} x_i + r)^2, \gamma > 0$
- Radial Basis Function (RBF)

$$K(x,x_i) = \exp\left(-\gamma ||x-x_i||^2\right), \gamma > 0$$

- Sigmoid $K(x,x_i) = \tanh(\gamma x^{TS} x_i + \gamma)$

We use the grid search to find which kernel is suited for our data set and achieved higher accuracy and f1-score.

## IV. RESULT ANALYSIS

The paper implements two different feature extraction techniques named Count Vector and TF-IDF. Features extraction techniques give us features into a numerical format that we can feed our classifiers Naïve Bayes classifiers: Multinomial Naïve Bayes and Support Vector Machine: binary classification separately. The evaluations are carried on using *Python*. *NumPy* [15, 16] and *Scikit-Learn* [17] are used to perform computations and implement machine learning models respectively.

To conduct the evaluation, the dataset was randomly split into training and testing set by a share of 70%-30%. We assumed a sentence as a document. We evaluate our classifiers with Count Vector and TF-IDF feature extraction techniques. During feature extraction, we use the default for the max features parameter of Count Vector and TF-IDF. We used 2000 sentence to extract features.

On the Naïve Bayes method, we use α=1 which symbolizes the Laplace Smoothing. We set the fit-prior-parameter=True on our Multinomial Naïve Bayes method to learn class prior probability that served more accuracy.

We used the grid search to find the optimal parameter for our support vector machine classifier with our dataset. We found that RBF kernel for C=100 and γ=0.01 give the maximum accuracy with count vector features. But on the TF-IDF feature SVM give maximum accuracy where C=1 and γ=1. Table IV describes the analysis of different classifiers with different feature extraction techniques.

TABLE IV. COMPARATIVE ANALYSIS WITH DIFFERENT CLASSIFIERS ON A DATASET

| Classifier | Feature Extraction Technique | Accuracy | Precision | Recall | F1 Score |
|---|---|---|---|---|---|
| Naïve Bayes Classifier | Count Vector | 0.885 | 0.91 | 0.89 | 0.90 |
|  | TF-IDf | 0.875 | 0.89 | 0.88 | 0.88 |
| SVM | Count Vector | 0.895 | 0.91 | 0.90 | 0.90 |
|  | TF-IDf | 0.89 | 0.95 | 0.89 | 0.92 |

We gained a maximum accuracy of 89% and F1-score of 92% with SVM classifier and TF-IDF feature extraction method. From Table IV, we can state that the difference in accuracy between the different techniques is more limited. But if we take attention to F1-score, we can conclude that SVM works better than the other classifier. However, accuracy can be increased by increasing the number of features. The number of features can be increased by collecting more data. We didn't use any stop word elimination process, which could a limit to receiving higher accuracy and F1-score on our dataset.

## V. CONCLUSION

Based on our research works presented in this paper, we may conclude that human abnormality can be classified from the Bengali text. We can classify abnormality of a person whether his/her expression is normal or abnormal by extracting his/her expressed text or converting his speech into text. We have acquired a good accuracy of around 89% on this study; however, this can be further enhanced. A rich dataset may help to gain much better accuracy by providing more information. As we have stated earlier that for the simplification and first attempt, we classified our attitude into only two classes in this research, we strongly believe that we will be able to extract more categorized sentiments in future research. We also believe that our contribution will inaugurate a wider perception in the expanse of human abnormality detection research works.

REFERENCES

[1] Neviarouskaya, Alena & Aono, Masaki & Prendinger, Helmut & Ishizuka, Mitsuru. (2014). Intelligent Interface for Textual Attitude Analysis. ACM Transactions on Intelligent Systems and Technology. 5. 1-20. 10.1145/2535912

[2] Karo Moilanen and Stephen Pulman, "Sentiment composition,", In Proceedings of the RANLP 2007, 378–382.

[3] Matthijs Mulder, Anton Nijholt, Marten den Uyl, and Peter Terpstra, "A lexical grammatical implementation of affect," In Proceedings of the TSD 2004, Springer, Berlin, 171–178.

[4] Tetsuya Nasukawa and Jeonghee Yi, "Sentiment analysis: Capturing favorability using natural language processing," In Proceedings of the K-CAP 2003, 70–77.

[5] A. Go, R. Bhayani and L. Huang. "Twitter sentiment classification using distant supervision". Technical report, Stanford Digital Library Technologies Project. 2009.

[6] D. Davidiv, O. Tsur and A. Rappoport, "Enhanced Sentiment Learning Using Twitter Hash-tags and Smileys". In Proceedings of the 23rd International Conference on Computational Linguistics: Posters, COLING '10, pp. 241–9. Stroudsburg, PA: Association for Computational Linguistics. 2010.

[7] Pak, A., & Paroubek, P. (2010, May). Twitter as a corpus for sentiment analysis and opinion mining. In LREc (Vol. 10, No. 2010, pp. 1320-1326).

[8] Balahur, A., & Turchi, M. (2012, July). Multilingual sentiment analysis using machine translation. In Proceedings of the 3rd Workshop in Computational Approaches to Subjectivity and Sentiment Analysis (pp. 52- 60). Association for Computational Linguistics.

[9] Kaur, A., & Gupta, V. (2014). Proposed algorithm of sentiment analysis for punjabi text. Journal of Emerging Technologies in Web Intelligence, 6(2), 180-183.

[10] Hasan, K. A., & Rahman, M. (2014, December). Sentiment detection from Bangla text using contextual valency analysis. In Computer and Information Technology (ICCIT), 2014 17th International Conference on (pp. 292-295). IEEE.

[11] Chowdhury, S., & Chowdhury, W. (2014, May). Performing sentiment analysis in Bangla microblog posts. In 2014 International Conference on Informatics, Electronics & Vision (ICIEV) (pp. 1-6). IEEE.

[12] Nabi, M. M., Altaf, M. T., & Ismail, S. (2016). Detecting sentiment from Bangla text using machine learning technique and feature analysis. International Journal of Computer Applications, 153(11).

[13] S. Chowdhury and W. Chowdhury, "Performing sentiment analysis in Bangla microblog posts," 2014 International Conference on Informatics, Electronics & Vision (ICIEV), Dhaka, 2014, pp. 1-6.

[14] N. Tabassum and M. I. Khan, "Design an Empirical Framework for Sentiment Analysis from Bangla Text using Machine Learning," 2019 International Conference on Electrical, Computer and Communication Engineering (ECCE), Cox'sBazar, Bangladesh, 2019, pp. 1-5.

[15] A. Neviarouskaya, H. Prendinger and M. Ishizuka, "@AM: Textual Attitude Analysis Model," Proceedings of the NAACL HLT 2010 Workshop on Computational Approaches to Analysis and Generation of Emotion in Text, pages 80–88, Los Angeles, California, June 2010. c 2010 Association for Computational Linguistics.

[16] Travis E. Oliphant. A guide to NumPy, USA: Trelgol Publishing, (2006).

[17] Stéfan van der Walt, S. Chris Colbert and Gaël Varoquaux. The NumPy Array: A Structure for Efficient Numerical Computation, Computing in Science & Engineering, 13, 22-30 (2011).

[18] Scikit-learn: Machine Learning in Python, Pedregosa et al., JMLR 12, pp. 2825-2830, 2011.